\documentclass[sigconf,natbib=true,anonymous=false]{acmart}
\renewcommand\footnotetextcopyrightpermission[1]{} 
\usepackage{bm}
\usepackage{booktabs}
\usepackage{multirow}
\usepackage{threeparttable}
\usepackage{adjustbox}
\usepackage[makeroom]{cancel}
\usepackage{amsmath}
\usepackage[ruled,vlined]{algorithm2e}
\usepackage{arydshln}
\usepackage{subcaption}
\AtBeginDocument{%
  \providecommand\BibTeX{{%
    \normalfont B\kern-0.5em{\scshape i\kern-0.25em b}\kern-0.8em\TeX}}}

\acmBooktitle{under review}

\begin{document}

\title{Physics-Coupled Spatio-Temporal Active Learning for Dynamical Systems}

\author{Yu Huang, Yufei Tang, Xingquan Zhu, Min Shi, Ali Muhamed Ali, Hanqi Zhuang, Laurent Cherubin}
\email{{yhwang2018, tangy, xzhu3, mshi2018, amuhamedali2014, zhuang, lcherubin}@fau.edu}
\affiliation{%
  \institution{Florida Atlantic University}
  \city{Boca Raton}
  \state{Florida}
  \country{USA}
  \postcode{33431}
}



\begin{abstract}
Spatio-temporal forecasting is of great importance in a wide range of dynamical systems applications from atmospheric science, to recent COVID-19 spread modeling. These applications rely on accurate predictions of spatio-temporal structured data reflecting real-world phenomena. A stunning characteristic is that the dynamical system is not only driven by some physics laws but also impacted by the localized factor in spatial and temporal regions. One of the major challenges is to infer the underlying causes, which generate the perceived data stream and propagate the involved causal dynamics through the distributed observing units. Another challenge is that the success of machine learning based predictive models requires massive annotated data for model training. However, the acquisition of high-quality annotated data is objectively manual and tedious as it needs a considerable amount of human intervention, making it infeasible in fields that require high levels of expertise. To tackle these challenges, we advocate a spatio-temporal physics-coupled neural networks (ST-PCNN) model to learn the underlying physics of the dynamical system and further couple the learned physics to assist the learning of the recurring dynamics. To deal with data-acquisition constraints, an active learning mechanism with Kriging for actively acquiring the most informative data is proposed for ST-PCNN training in a partially observable environment. Our experiments on both synthetic and real-world datasets exhibit that the proposed ST-PCNN with active learning converges to near optimal accuracy with substantially fewer instances.
\end{abstract}





\maketitle

\section{Introduction}
Spatio-temporal modeling is essential in many scientific fields ranging from studies in biology \cite{li2019regional}, information flow in social networks \cite{li2019learning}, traffic predictions \cite{nguyen2016discovering}, and atmospheric science \cite{liu2019nonparametric}, to recent COVID-19 spread modeling \cite{la2020epidemiological}. These applications rely on accurate predictions of spatio-temporal structured data reflecting real-world phenomena. With an unprecedented increase in data accessibility and computational capability, machine learning (ML) recently made a breakthrough and has been rapidly explored in spatio-temporal modeling \cite{wang2020calendar,dai2020hybrid,li2020autost,wang2020calendar,yu2017spatio,seo2019differentiable}.

One of the major challenges of spatio-temporal ML is the high cost of data acquisition. Effective training of advanced ML models requires large amounts of \textit{labeled} data. However, the acquisition of a large number of high-quality annotated data consumes a lot of workforces, making it unfeasible in fields that require high levels of expertise \cite{ren2020survey}. This problem is outstanding, especially in geographic and atmospheric science, which involves the accurate prediction of some variables of interest \cite{zhao2018correlation,lee2019deeproof}. 
For example, collecting ocean data usually requires trained scientists to travel to sampling locations and deploy in-situ instruments within a stream. The cost further limits the total number of sensors placed/installed to query the data; while the incurring personnel and equipment costs for data does not necessarily mean improved model predictions.

An elementary question for spatio-temporal forecasting is where and when to query the data within constraints to build efficient and accurate predictive models. Active learning (AL) is a special form of weakly supervised learning, attempting to maximize a model’s performance gain while annotating the fewest samples possible. Therefore, it is natural to investigate whether AL can be used to reduce the cost of sample annotations in the spatio-temporal domain \cite{ren2020survey}. AL approaches, from application scenarios \cite{settles2009active}, can be categorized into membership query synthesis \cite{king2004functional}, stream-based selective sampling \cite{krishnamurthy2002algorithms}, and pool-based AL \cite{tsiakmaki2020fuzzy}. Among these approaches, stream-based selective sampling makes an independent judgment on whether each sample in the data stream needs to query unlabeled samples' labels, and is more suitable for spatio-temporal scenarios since timeliness is required and the whole dataset is not always available.

Another major challenge of spatio-temporal learning for dynamical systems is to infer the underlying causes, which generate the perceived data stream and propagate the involved causal dynamics through distributed sensor meshes. A stunning characteristic of such scenarios is that the widely distributed sensors share striking \textit{homogeneity} and \textit{heterogeneity}. The former is driven by the physical laws governing the systems, whereas the latter is impacted by the localized factor in spatial and temporal regions. However, existing ML methods seldom include prior knowledge of the underlying physics, i.e., considering the \textit{homogeneity}. A critical property that all spatio-temporal processes have in common is that some general underlying principles will apply irrespective of time or location when observing natural processes. For example, many climate models based on mathematical equations that describe the physical processes have been built to predict climate variables while the prediction abilities may vary dramatically across other regions and time \cite{liu2019nonparametric}. Although physics-informed ML has emerged to build hybrid models for robust predictions recently \cite{rudy2017data,berg2019data,xu2019dl,long2018pde,hu2020revealing}, these methods neither consider the homogeneity and heterogeneity simultaneously nor the data limitation.

To address these challenges, we introduce a spatio-temporal physics-coupled neural networks (ST-PCNN) model with active learning to capture the spatio-temporal correlations, heterogeneity, and its inherited homogeneity in a spatially distributed manner with a limited data source. The main contributions of this paper are the following:
\begin{itemize}
     \item We propose an active learning algorithm for spatial-temporal dynamical systems, where Kriging sampling is designed to obtain the critical training data at locations where underlying predictions are in most need of improvement.
    \item We propose ST-PCNN, a novel framework for spatio-temporal modeling. It learns a forecasting neural network distributively executed at different locations, integrating lateral information interacting with the neighborhoods. ST-PCNN allows efficient parallel computation and captures heterogeneity from all spatial locations.
    \item We also propose embedding a physics learning module into the ST-PCNN to learn the inherited homogeneity (e.g., in the form of partial differential equation (PDE)). The physics module first fits a Gaussian process (GP) to the observations and then obtains the PDE coefficients by calculating the GP-derived partial derivatives of the state variable, without the need for initial and boundary conditions.
    \item We conduct extensive experiments and comparative studies on both synthetic and real-world datasets, where the former dataset manifests a clear spatio-temporal correlation and the latter presents strong sparsity.
\end{itemize}

\begin{figure*}[t]
\begin{small}
    \centering
    \includegraphics[width=\textwidth]{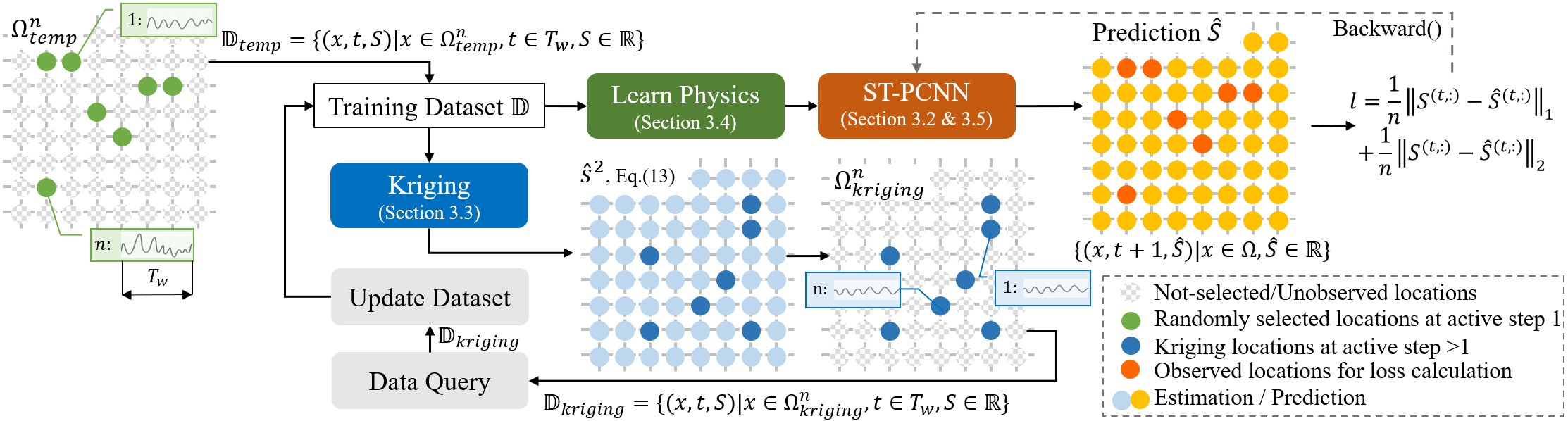}
    \caption{The overall framework of the proposed ST-PCNN model. A system consists of $N$ observable locations, but can only collect values from $n$ locations at any particular time. In the beginning, a set of $n$ random locations are selected (the green-colored dots on the top-left panel) to collect the training dataset $\mathbb{D}$. The lower panel loop denotes Kriging-based active learning, which actively queries locations at the next step, and uses queried samples to update the training set $\mathbb{D}$. Using the updated training dataset, the ``Learn Physics'' module learns the underlying physics, and combines it with a forecasting network to predict future values of all $N$ observable locations (the orange-colored dots on the top right). Because the red-colored dots contain queried values, they are used for loss calculation.} 
    \label{fig:active}
\end{small}
\end{figure*}

\section{Problem Definition}
An observing system (e.g., ocean observation sensor network \cite{hamilton2016loop}) consists of $N$ observable locations ($|\Omega|$=$N$), each produces a set of real valued observations in a temporal order. Due to resource constraints, such as timeliness or communication bandwidth, at any particular data collection period, the system can only collect values from $n$ locations ($0 < n \ll N$), denoted as $\Omega^n_{temp}$. The collected training set at each period is $\mathbb{D}_{temp} = \{(\bm{x}, t, \mathcal{S})|\bm{x}\in \Omega^n_{temp}, t\in T_w, \mathcal{S}\in\mathbb{R}\}$, where $\mathcal{S}$ represents the observation value, $\bm{x}$ is the location, and $T_w$ is the maximum length of up-to-date data stored in each location due to memory capacity.

Given a dynamical system with $N$ observable locations and their values up to current time $t$, and a budget $n$ restricting the number of query locations, our research has a \textbf{twofold-goal} to (1) find optimal $n$ locations to query their values; and (2) create a model to forecast future values of the \textit{whole system}, beyond the current time point (\textit{w.r.t} $t+1, t+2, \cdots$), with minimum forecasting errors.

To achieve the goal, we propose a framework with two main components: (1) Kriging based active learning to query high uncertainty points in the system; and (2) physics-coupled spatial-temporal learning to combine observed values and learned physics of the system to predict future values. Therefore, 
at current time point $t$, our system has observed $\mathbb{D}_{temp}$ of previous period (\textit{i.e.}, $t-T_w$ to $t$) from selected sites, the active learning module will determine where to query $n$ sites in the next period for best possible predictions. Because such queried values do not guarantee minimum forecasting errors, the physics-informed learning is introduced to leverage observed values to learn physics, and predict future values. The query and forecasting are essentially related, so our proposed framework delivers a closed-loop solution to ensure that active learning query and dynamics forecasting are combined to achieve the designed optimization goal.  

\section{Method}
\subsection{Spatio-Temporal Active Learning}
To model the dynamical systems considering both where and when to measure and query the data for making the best possible predictions while staying within a maximum budget of sensors and data, a spatio-temporal physics-coupled neural networks (ST-PCNN) model with active learning is proposed.

Active learning is a paradigm in which the network training procedure identifies and requests additional, high-benefit training data from an \textit{oracle}. In our case, the \textit{oracle} is an observing grid $\Omega$ constrained by the maximal data collection cost. The goal is to train the model to accurately predict the system dynamics everywhere within desired area/grid $\Omega$, using training data from a minimal number of locations in $\Omega_{temp}$. Our algorithm falls into the category of stream-based selective sampling active learning, proceeds by Algorithm \ref{algorithm:active} and illustrated in Figure \ref{fig:active}.

\begin{algorithm}[h]
\begin{small}
\SetAlgoLined
\SetKwInOut{Input}{Input}
\SetKwInOut{Output}{Output}
\SetKwBlock{Initialize}{Initialize}{}
\Input{(1) $\Omega$, a dynamical system as a grid; (2) $n$, budget restricting maximum \# of query locations at each time $t$\;}
\Output{ST-PCNN, the trained spatio-temporal model}
\Initialize{$\Omega^n_{temp}\leftarrow$~Randomly select $n$ locations from $\Omega$ to form a temporal observation subset\; $\mathbb{D} = \mathbb{D}_{temp}\leftarrow$~Collect $T_w$ consecutive observations from each selected locations to form the training data\; [\textit{Termination~ Conditions}]$\leftarrow$~ Threshold of prediction loss $\ell$ to terminate active learning;}
\While{Termination Conditions NOT satisfied}{
\textbf{Learn Physics}: $\left[\bm{\lambda} \right]\leftarrow$~ learn the physics, \textit{i.e.} coefficients $\bm{\lambda}$ of PDE, from the existing training data $\mathbb{D}$ by minimizing Eq. (\ref{Eq.ssre})\;
\textbf{ST-PCNN Training}: train the ST-PCNN with the learned physics $\lambda$ from training data $\mathbb{D}$, see Algorithm \ref{algorithm:ST-PCNN}\;
\textbf{Prediction}:$\left[\bm{\hat{\mathcal{S}}}\right]\leftarrow$ make prediction at all locations in the network, $\forall\bm{x}\in \Omega$\;
\textbf{Kriging}: $\Omega^n_{kriging}\leftarrow$~use Kriging to identify $n$ query locations with the largest estimated errors for next active learning step\;
\textbf{Data Query}: obtain consecutive observations of window size $T_w$ from the newly selected locations $\mathbb{D}_{kriging} = \{(\bm{x}, t, \mathcal{S})|\bm{x}\in \Omega^n_{kriging}, t\in T_w, \mathcal{S}\in\mathbb{R}\}$\; 
\textbf{Update Dataset}: add $\mathbb{D}_{kriging}$ to the existing training data $\mathbb{D}$\;
}
\caption{Active Learning for ST-PCNN}
\label{algorithm:active}
\end{small}
\vspace{-1mm}
\end{algorithm}

\begin{figure}[t]
\begin{small}
    \centering
    \includegraphics[width=0.49\textwidth]{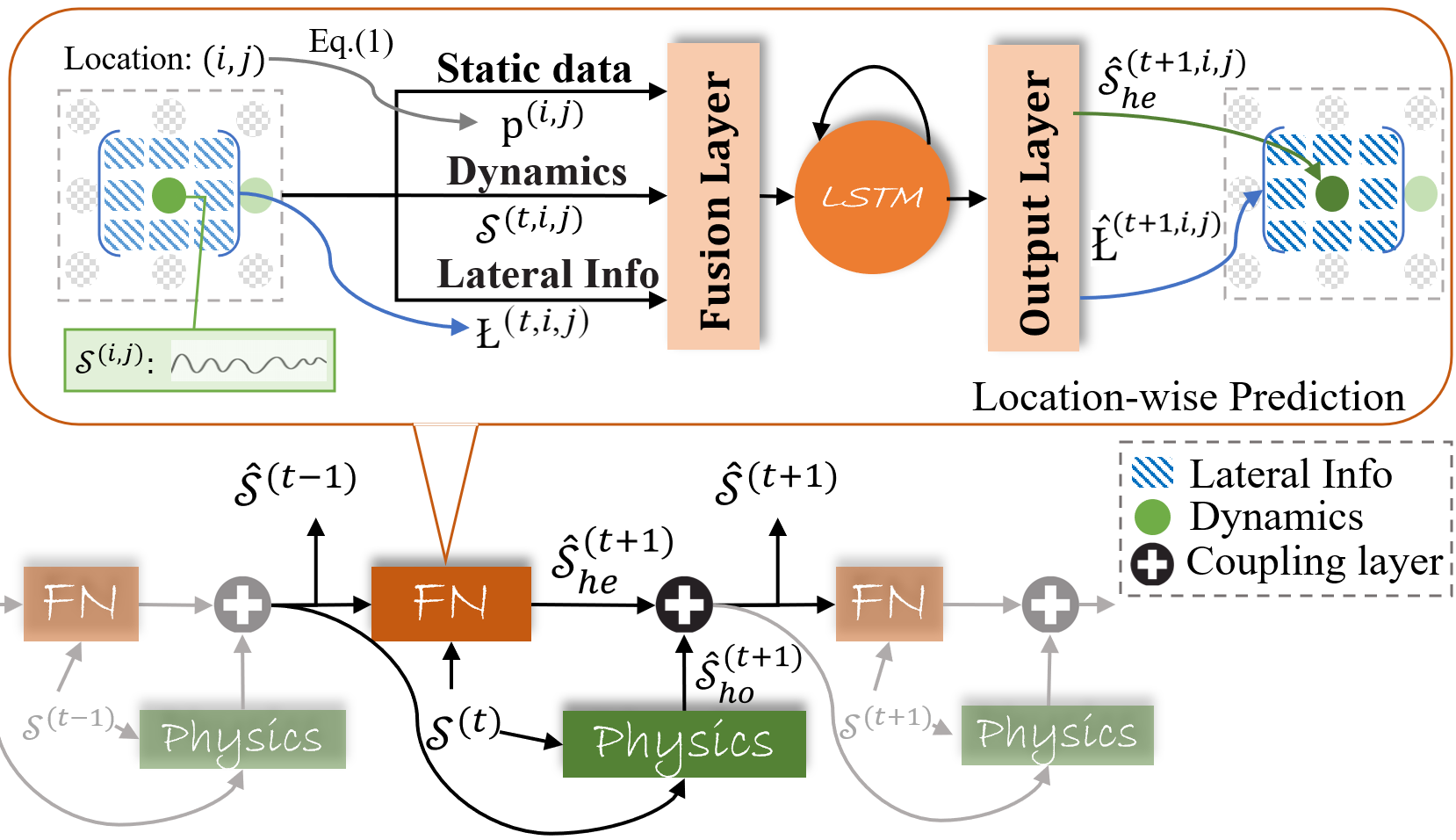}
    \caption{The lateral connection schema of forecasting network (FN) and physics network (PN). The upper panel unfolds the structure of the FN network. The green node denotes a center node located at $(i,j)$, the grey (unobserved) and light green (observed) nodes are its neighbors. Three types of information are used to characterize each node: (1) $\mathcal{S}^{(t,i,j)}$: an embedding vector representing dynamics of node at time $t$, (2) $\mathbf{\vec{p}}^{(i,j)}$: an embedding vector representing node location; and (3) $\L^{(t,i,j)}$: an embedding vector (dashed dot-square set) capturing interaction (lateral info) between each node and its neighbors.}
    \label{fig:connection}
\end{small}
\end{figure}


\subsection{ST-PCNN Framework}
ST-PCNN is a bi-network architecture, as shown in Figure \ref{fig:connection}, including a physics network (PN) and a forecasting network (FN). The FN produces \textit{heterogeneous} prediction leveraging its own specific local attributes only, while the PN generates \textit{homogeneous} solution of the dynamics regularized by the overall underlain governing physics.

FN receives 1) dynamic data, that changes over time, 2) static information, that stays constant and characterizes the location of each FN, and 3) lateral information from neighbors. The output of each FN includes predicted dynamics and additional lateral information denoting interaction between neighbors. Such interaction, which distinguishes our architecture from others, aims to model the location-sensitive transitions between adjacent FNs and thus enable local context-dependent spatial information propagation.


In many natural phenomenons, data are collected distributively and exhibit heterogeneous properties: each of these distributed locations presents a different view of the natural process at the same time, where each view has its own individual representation space and dynamics. Theoretically, each location may contain information that other locations do not have access to. Therefore, all local views must interact in some way in order to describe the global activity comprehensively and accurately.

How to explicitly encode location information into neural networks is critical in our location-wise forecasting. Inspired by the Transformer that encodes word positions in sentences, we extend the absolute positional encoding to represent grid positions. In particular, let $i,j$ be the desired position in a regular grid, $\mathbf{\vec{p}}^{(i,j)}\in\mathbb{R}^d$ be its corresponding encoding, and $d$ be the encoding dimension, then the encoding scheme is defined as:
\vspace{-2mm}
\begin{equation}
\mathbf{\vec{p}}^{(i,j)} :=
\left\{\begin{array}{ll}
\mathrm{sin}(\omega_k, i), \mathrm{sin}(\omega_k, j) & \mathrm{if}\;i,j=2k \\ 
\mathrm{cos}(\omega_k, i), \mathrm{cos}(\omega_k, j) & \mathrm{if}\;i,j=2k+1
\end{array}\right.
\vspace{-2mm}
\end{equation}
where $\omega_k =\frac{1}{10,000^{2k/d}}$, $k\in\mathbb{N}_{\leq \left\lceil\frac{d}{2}\right\rceil}$. The positional embedding as a vector contains pairs of $sines$ and $cosines$ for each decreasing frequency along the vector dimension.

As illustrated in Figure \ref{fig:connection}, at each time $t$, FN is distributively executed at different locations $\Omega_{temp}^n$. FN encodes each view (i.e., static $\mathbf{\vec{p}}$, dynamics $\mathcal{S}$, and $\L$ of each node) using a fusion layer: 
\begin{equation}
    f^{(t,i,j)} = [\mathbf{\vec{p}}^{(i,j)}, \mathcal{S}^{(t,i,j)},\L^{(t,i,j)}]\mathcal{W}_{fusion}^T + b_{fusion} 
\end{equation}
where $\L^{(t,i,j)}$ is a vector used to characterize interaction between node at $i,j$ and its neighbors. It is initialized as zeros at the first step and continuously updated by Eq.~(\ref{Equ:fn_out}) when $t>0$.

These features, $f^{(t,i,j)} \in \mathbb{R}^{d_{f_{i,j}}}$, are then fed into an LSTM to model the node-specific interactions over time. The update mechanism of the LSTM cell is defined as:
\begin{equation}
\label{Equ:lstm}
    \left [\mathcal{I}^{(t)};\mathcal{F}^{(t)};\tilde{\mathcal{C}}^{(t)};\mathcal{O}^{(t)} \right ] = \sigma\left ( \mathcal{W} \cdot f^{(t,i,j)} + \mathcal{T} \cdot h^{(t-1)} \right )
\end{equation}
\begin{equation}
    \mathcal{C}^{(t)}=\tilde{\mathcal{C}}^{(t)}\circ \mathcal{I}^{(t)}; h^{(t)}=\mathcal{O}^{(t)}\circ\mathcal{C}^{(t)}
\end{equation}
where $\sigma(\cdot)$ applies sigmoid on the input gate $\mathcal{I}^{(t)}$, forget gate $\mathcal{F}^{(t)}$, and output gate $\mathcal{O}^{(t)}$, and $tanh(\cdot)$ on memory cell $\tilde{\mathcal{C}}^{(t)}$. The parameters are characterized by $\mathcal{W}\in \mathbb{R}^{d_{f_{i,j}}\times d_{h_{i,j}}}$ and $\mathcal{T}\in\mathbb{R}^{d_{h_{i,j}}\times d_{h_{i,j}}}$, where $d_{h_{i,j}}$ is the output dimension. A cell updates its hidden states $h^{(t)}$ based on the previous step $h^{(t-1)}$ and the current input $f^{(t,i,j)}$.

An output layer is stacked at the end of FN to transform the LSTM output into the expected dynamic prediction and additional lateral information as:
\begin{equation}\label{Equ:fn_out}
    \left [\hat{\mathcal{S}}_{he}^{(t+1,i,j)};\hat{\L}^{(t+1,i,j)} \right ] = Relu(\mathcal{W}_{out}\cdot f^{(t,i,j)} + b_{out}) 
\end{equation}
where $\hat{\mathcal{S}}_{he}^{(t+1,i,j)}$ denotes the \textit{heterogeneous} prediction of the node dynamics at time step $t+1$. The learnable parameters are characterized by $\mathcal{W}_{out}\in\mathbb{R}^{d_{f_{i,j}}\times d_{y_{i,j}}}$ and $b_{out}\in\mathbb{R}^{d_{y_{i,j}}}$, where $d_{y_{i,j}}$ denotes the total dimensions of the dynamic and the lateral outputs.

\subsection{Kriging Sampling for Active Learning}
Kriging, also know as Gaussian process regression, is a widely used method in applied mathematics and machine learning for constructing surrogate models, interpolation, supervised learning, and active learning \cite{faghihpirayesh2020motor}. Kriging constructs a statistical model of a partially observed function of time and/or space, assuming that this function is a realization of a Gaussian process (GP) that is uniquely described by its mean and covariance.

Denote the observation locations as $\Omega=\{\bm{x}^{(i)}\}_{i=1}^N$, where $\bm{x}^{(i)}$ are $d$-dimensional vectors in $\mathcal{D}\in \mathbb{R}^d$ and the observed state values at these locations as $\bm{y}=\{y^{(i)}\}_{i=1}^N$, where $y^{(i)}\in\mathbb{R}$. The Kriging assumes that the observation vector $\bm{y}$ is a realization of the following N-dimensional random vector that satisfies multivariate Gaussian distribution:
\begin{equation}
    \mathcal{Y} = \left( \mathcal{Y}(\bm{x}^{(1)}), \mathcal{Y}(\bm{x}^{(2)}), \cdots, \mathcal{Y}(\bm{x}^{(N)})\right)^T
\end{equation}
where $\mathcal{Y}(\bm{x}^{(i)})$ is the concise notation of $\mathcal{Y}(\bm{x}^{(i)};\omega)$, which is a Gaussian random variable defined on a probability space $(\Lambda, \mathcal{E}, \mathcal{P})$ with $\omega \in\Lambda$, where $\Lambda$ is the sample space, $\mathcal{E}$ is the $\sigma$-algebra over $\Lambda$, and $\mathcal{P}$ is the probability measure on $\Lambda$ \cite{yang2018physics}. Of note, $\bm{x}(i)$ can be considered as the parameters of the GP $\mathcal{Y}: \mathcal{D}\times\Lambda \rightarrow \mathbb{R}$, such that $\mathcal{Y}(\bm{x}^{(i)}):\Lambda\rightarrow\mathbb{R}$ is a Gaussian random variable for any $\bm{x}^{(i)}$ in the set $\mathcal{D}$. Usually, $\mathcal{Y}(\bm{x})$ is denoted as:
\begin{equation}
    \mathcal{Y}(\bm{x}) \sim \mathcal{GP}(\mu(\bm{x}),k(\bm{x},{\bm{x}}'))
\end{equation}
where $\mu: \mathcal{D}\rightarrow\mathbb{R}$ is the mean, $k(\bm{x},{\bm{x}}'):\mathcal{D}\times\mathcal{D}\rightarrow \mathbb{R}$ is the covariance function, also known as kernel, that provides a measure of closeness between a training point $\bm{x}$ and a test point ${\bm{x}}'$:
\begin{equation}
    \mu(\bm{x}) = \mathbb{E}\{\mathcal{Y}(\bm{x})\}
\end{equation}
\begin{equation}
      k(\bm{x}, {\bm{x}}') = \mathbb{E}\{(\mathcal{Y}(\bm{x})-\mu(\bm{x}))(\mathcal{Y}({\bm{x}}')-\mu({\bm{x}}'))\}  
\end{equation}

The variance of $\mathcal{Y}(\bm{x})$ is $k(\bm{x}, \bm{x})$ and its standard deviation is $\sigma(\bm{x}) = \sqrt{k(\bm{x},\bm{x})}$. The covariance matrix $\mathcal{C}$ of a random vector $\mathcal{Y}$ is defined as:
\begin{equation}
    \mathcal{C} = \begin{bmatrix}
k(\bm{x}^{(1)}, \bm{x}^{(1)}) & \cdots & k(\bm{x}^{(1)}, \bm{x}^{(N)}) \\ 
\vdots  & \ddots & \vdots  \\ 
k(\bm{x}^{(N)}, \bm{x}^{(1)}) & \cdots & k(\bm{x}^{(N)}, \bm{x}^{(N)}) 
\end{bmatrix}
\end{equation}
then, the estimation at any new location $x^{*}$ is given as:
\begin{equation}
    \hat{y}(\bm{x}^{*}) = \mu(\bm{x}^{*}) + \bm{c}^T\mathcal{C}^{-1}(\bm{y}-\bm{\mu})
\end{equation}
where $\bm{\mu} = \left[ \mu(\bm{x}^{(1)}),\cdots,\mu(\bm{x}^{(N)} )\right]^T$ and $\bm{c}$ is a vector of covariance between the observed data and the predictions:
\begin{equation}
    \bm{c} = c({\bm{x}}^{*})=\left[k(\bm{x}^{(1)}, \bm{x}^{*}), k(\bm{x}^{(2)}, \bm{x}^{*}), \cdots, k(\bm{x}^{(N)}, \bm{x}^{*})\right]^T
\end{equation}

The mean squared error (MSE) of this prediction is defined as $\hat{s}^2(\bm{x}^{*})=\mathbb{E}\{(\hat{y}(\bm{x}^*)-\mathcal{Y}(\bm{x}^*))^2\}$ and is calculated as 
\begin{equation}
    \hat{s}^2(\bm{x}^{*}) = \sigma^2(\bm{x}^{*}) - \bm{c}^T\mathcal{C}^{-1}\bm{c}
\end{equation}

The prediction and MSE can be derived from the maximum likelihood estimate (MLE) method. The $n$ points with the largest MSE values are selected as the next query points. This policy reassembles to the uncertainty sampling in active learning, so the most uncertain points are selected for future query. 

In this context, the Kriging sampling is a process of identifying new locations for critical observations that minimize the prediction error and reduce MSE or uncertainty. Notably, Algorithm \ref{algorithm:active}  is a greedy algorithm to identify observation locations when some observations are affordable. It does not guarantee to identify the \textit{actual optimal} new observation locations, but the \textit{estimated optimal}. Our active learning framework, unlike others that keep adding new locations for data collection, is constrained by maximal data collection cost that only a fixed limited number of locations are available at each period. A natural way to improve the predictive performance is to consider leveraging observations at the locations corresponding to local maximum in $\hat{s}^2(\bm{x}^{*})$. Then, we can train our model on those selected locations to make a more accurate prediction and compute a new $\hat{s}^2(\bm{x}^{*})$ to select the locations for the next period's data collection, training, and prediction.

\subsection{Physics Learning}
Physicists attempt to model natural phenomena in a principled way through analytic descriptions. Conservation laws, physical principles, or phenomenological behaviors are generally formalized using differential equations, which can best derive the homogeneity of observations. The knowledge accumulated for modeling physical processes in well-developed fields such as maths or physics could be a useful guideline for dynamics learning. ST-PCNN includes a physics-aware module (\textit{i.e.}, PN) to learn underlying hidden physics. The objective of the PN is to estimate parameters of a partial differential equation (PDE), given observations of state variable.

The PN requires fitting a GP model to the observations of the state variable. The derivatives of the state variable can be obtained using the property that 'the derivative of a GP is also a GP' \cite{solak2003derivative}, and finally adjusting the PDE coefficients so that the GP derived partial derivatives satisfy the PDE \cite{rai2019gaussian}. This module does not require initial and boundary conditions.

The Gaussian covariance function, that notationally given by $k(\bm{v},{\bm{v}}')$, also known as kernel, is:
\begin{equation}
    k(\bm{v},{\bm{v}}') = \sigma_{s}^2exp\left( -\frac{1}{2}\sum_{j=1}^{m}\frac{(v_j-{v}'_j)^2}{l_j^2}\right)
\end{equation}
where $v$ is the collection of variables $\{\bm{x}, t\}$, $m$ is the dimension of the variables space, \textit{i.e.}, the number of independent variables in $\bm{v}$ and $v_j\in\bm{v}$, $l_j$ is the length scale for the $j$-th independent variable $v_j$, and $\sigma_s$ is the scale parameter of the covariance function. Together, they are referred as hyperparameters and are represented by $\bm{\phi} = [l_1, \cdots, l_m, \sigma_s]$. The Gaussian covariance function, also known as squared exponential covariance function, is infinitely differentiable and hence very smooth. It is probably the most widely used covariance function in GP regression models.

Let the dataset be $\mathbb{D} = \{(\bm{v}_i, y_i)|i=1,\cdots,N;\bm{v}_i\in \mathbb{R}^m; y_i\in \mathbb{R}\}$, where $N$ is the number of observations, $\bm{v}_i$ is an input vector (independent variable) in m-dimensional space, and $y_i$ is the corresponding observed value of the dependent variable. Also assume that all of the $y_i$ have independent and identically distributed noise, which is assumed to be normally distributed with mean zero and standard deviation $\sigma_y$:
\begin{equation}
    y_i = \tilde{y}_i + \mathcal{N}(0, \sigma_y^2)
\end{equation}
where $\tilde{y}_i$ is the theoretical value of the state variable $\bm{v}_i=[\bm{x}_i, t]$ at location $\bm{x}_i$ and time $t$. Suppose $\bm{y}$ follows a Gaussian distribution with mean $\bar{\bm{\mathcal{V}}}$ and covariance matrix $\mathcal{C}$, represented as:
\begin{equation}
    \bm{y} \sim \mathcal{N}(\bar{\bm{\mathcal{V}}}, \mathcal{C})
\end{equation}
where $\bar{\bm{\mathcal{V}}}$ is a $N\times1$ vector of the priori mean of GP at data points $\mathbb{D}$. In most cases the a priori mean may not be available, but the GP model is powerful enough to capture trends in the data even if the mean of GP is taken as zero or a constant value.

The estimation $\bm{y}^*$ at new points $\bm{v}^* \in \bm{\mathcal{V}}^*$ is given by:
\begin{equation}
    \bm{y}^{*} \sim \mathcal{N}(\bar{\bm{\mathcal{V}}^*}, \mathcal{C}^{**})
\end{equation}
where $\mathcal{C}^{**} = k(\bm{v}^*,\bm{v}^*)$. The joint distribution of $\bm{y}$ and $\bm{y}^{*}$ is given by:
\begin{equation}
    \begin{bmatrix} \bm{y}\\ \bm{y}^* \end{bmatrix} = \mathcal{N}\left( \begin{bmatrix}\bar{\bm{\mathcal{V}}}\\ \bar{\bm{\mathcal{V}}^*} \end{bmatrix},\begin{bmatrix}
    \mathcal{C} + \sigma_y^2\bm{I} & \mathcal{C}^* \\ {\mathcal{C}^*}^T & {\mathcal{C}^{**}}
    \end{bmatrix}\right)
\end{equation}
where $\bm{I}$ is the identity matrix, $\mathcal{C}^*=k(\bm{v},\bm{v}^*)$ and the elements of $\mathcal{C}^*$ are covariance between the $i$th observed point and the $j$th new point (i.e., $k_{ij}$).

The predictive distribution $p(\bm{y}^*|\bm{V}^*,\bm{\phi},\sigma_y,\bm{V}, \bm{y})$ based on the conditional property of the Gaussian distribution is given by:
\begin{equation}
    p(\bm{y}^*|\bm{V}^*,\bm{\phi},\sigma_y,\bm{V}, \bm{y}) = \mathcal{N}(\bm{y}^*|\bm{\mu}, \bm{\Sigma})
\end{equation}
where
\begin{equation}\label{Eq:mu}
\bm{\mu} = \mathbb{E}[\bm{y}^*|\bm{V}^*,\bm{\phi},\sigma_y,\bm{V}, \bm{y}]= \bar{\bm{\mathcal{V}}^*} + {\mathcal{C}^*}^T(\mathcal{C}+\sigma_y^2\bm{I})^{-1}(\bm{y}-\bar{\bm{\mathcal{V}}})
\end{equation}
\begin{equation}
\bm{\Sigma} = cov[\bm{y}^*|\bm{V}^*,\bm{\phi},\sigma_y,\bm{V}, \bm{y}] =  \mathcal{C}^{**} - {\mathcal{C}^*}^T(\mathcal{C}+\sigma_y^2\bm{I})^{-1}\mathcal{C}^* + \sigma_y^2\bm{I}
\end{equation}

Estimating the PDE coefficients requires derivatives of the state variable (fitted by a GP model) with respect to independent variables. According to \cite{solak2003derivative}, the GP relation Eq. (\ref{Eq:mu}) could be straightforwardly differentiated with respect to input variable. The first-derivatives of the covariance function is presented as:
\begin{equation}
        \frac{\partial k(\bm{v},\bm{v}*)}{\partial v^*_j} 
      = \frac{\partial}{\partial v^*_j}\left[\sigma_s^2exp\left( -\frac{1}{2}\sum_{i=1}^m \frac{(v_i-v^*_i)^2}{l_i^2}\right)\right] = k(\bm{v},\bm{v}*)\frac{(v_j-v^*_j)}{l_j^2}
\end{equation}
The second-derivatives of the covariance function is presented as:
\begin{equation}
        \frac{\partial^2 k(\bm{v},\bm{v}*)}{\partial {v^*_j}^2}
         = \frac{\partial}{\partial v^*_j}\left[k(\bm{v},\bm{v}*)\frac{(v_j-v^*_j)}{l_j^2}\right]
        =\frac{k(\bm{v},\bm{v}*)}{l_j^2}\left[\frac{(v_j-v^*_j)^2}{l_j^2}-1\right]
\end{equation}

\begin{equation}
        \frac{\partial^2 k(\bm{v},\bm{v}*)}{\partial {v_jv^*_j}}
         = \frac{\partial}{\partial v^*_j}\left[k(\bm{v},\bm{v}*)\frac{(v_j-v^*_j)}{l_j^2}\right]=\frac{k(\bm{v},\bm{v}*)}{l_j^2}\left[1-\frac{(v_j-v^*_j)^2}{l_j^2}\right]
\end{equation}

For a constant mean GP, the first order derivative $\dot{\bm{y}}_{gp}$ is given by $p({\dot{\bm{y}}}_{gp}|\bm{y}, \bm{\phi})=\mathcal{N}(\dot{\bm{\mu}}, \dot{\bm{\Sigma}}_{gp})$, where $\dot{\bm{\Sigma}}_{gp}$ is the covariance matrix of $\dot{\bm{y}}_{gp}$. Specifically:
\begin{equation}
\dot{\bm{\mu}} = \frac{\partial}{\partial v^*_j}\mathbb{E}[\bm{y}^*|\bm{V}^*,\bm{\phi},\sigma_y,\bm{V}, \bm{y}] = \dot{{\mathcal{C}}^*}^T(\mathcal{C}+\sigma_y^2\bm{I})^{-1}(\bm{y}-\bar{\bm{\mathcal{V}}})
\end{equation}
\begin{equation}
    \dot{\bm{\Sigma}}_{gp} = {\dot{\mathcal{C}}}^{**} - {\dot{\mathcal{C}^*}}^T(\mathcal{C}+\sigma_y^2\bm{I})^{-1}{\dot{\mathcal{C}}}^*
\end{equation}
where $\dot{\mathcal{C}^*}$ is populated by the mixed covariance function between the state variable and its first order partial derivatives, $cov(\bm{y}, \frac{\partial}{\partial v^*_j}\bm{y}^*) = \frac{\partial}{\partial v^*_j}k(\bm{v}, \bm{v}^*)$, also known as cross-covariance between the state variable and its derivative. ${\dot{\mathcal{C}}}^{**}$ is a mixed
covariance function between partial derivatives:
\begin{equation}
cov(\frac{\partial}{\partial v_j}\bm{y}, \frac{\partial}{\partial v^*_j}\bm{y}^*) = \frac{\partial^2}{\partial v_j \partial v^*_j}k(\bm{v}, \bm{v}^*)   
\end{equation}

Accordingly, the second-derivative of $\mathcal{GP}$ is given by:
\begin{equation}
\ddot{\bm{y}}_{gp} = \frac{\partial^2}{\partial {v^*_j}^2}\mathbb{E}[\bm{y}^*|\bm{V}^*,\bm{\phi},\sigma_y,\bm{V}, \bm{y}]  = \ddot{{\mathcal{C}}^*}^T(\mathcal{C}+\sigma_y^2\bm{I})^{-1}(\bm{y}-\bar{\bm{\mathcal{V}}})
\end{equation}
where $\ddot{{\mathcal{C}}^*}^T$ is populated by the mixed covariance function between the state variable and its second order partial derivative:
\begin{equation}
cov(y, \frac{\partial^2}{\partial {v^*_j}^2}\bm{y}^*) = \frac{\partial^2}{\partial {v^*_j}^2}k(\bm{v}, \bm{v}^*)   
\end{equation}

Let the hidden PDE be given by an implicit function as:
\begin{equation}
f( v_1, \cdots, v_m, y, \frac{\partial y}{\partial v_1},\cdots, \frac{\partial y}{\partial v_m}, \frac{\partial^2 y}{\partial v_1\partial v_1}, \cdots, \frac{\partial^2 y}{\partial v_1\partial v_m}, \cdots, \lambda )
\end{equation}
where $y$ is the state variable, $v_1, \cdots, v_m$ are independent variables, and $\lambda$ is a set of PDE coefficients. By given the observations of the state variable and its GP derivatives involved in the PDE, the residual error in the PDE at an observed point is given by:
\begin{equation}
\begin{split}
    \epsilon = f\bigg(&  v_1, \cdots, v_m, y, \lambda\\
     & \mathcal{GP}\{\frac{\partial y}{\partial v_1},\cdots, \frac{\partial y}{\partial v_m}, \frac{\partial^2 y}{\partial v_1\partial v_1}, \cdots, \frac{\partial^2 y}{\partial v_1\partial v_m}, \cdots,\}\bigg)  
\end{split}
\end{equation}
where the notation $\mathcal{GP}(\cdot)$ incorporates all the partial derivatives of the state variable, means that the arguments are evaluated using the GP relation. The coefficients set $\lambda$ can be obtained by assuming a distribution for the residual errors and then applying either the method of maximum likelihood or the Bayesian estimation method using Markov chain Monte Carlo sampling. These methods not only provide a point estimate of $k$ but also an estimation interval. However, to keep the parameter estimation simple, we have obtained $k$ by minimizing the sum of square of residual error (SSRE), given as:
\begin{equation}\label{Eq.ssre}
\begin{split}
     SSRE & = \epsilon^T\epsilon  = \sum_{\forall \bm{v}, \bm{y} \in \mathbb{D}} f\bigg(v_1, \cdots, v_m, y, \lambda,\\ 
     & \mathcal{GP}\{\frac{\partial y}{\partial v_1},\cdots, \frac{\partial y}{\partial v_m}, \frac{\partial^2 y}{\partial v_1\partial v_1}, \cdots, \frac{\partial^2 y}{\partial v_1\partial v_m}, \cdots,\}\bigg)^2
\end{split}
\end{equation}

The minimization is performed using Nelder-Mead algorithm. The GP regression model provides standard error for the estimated state variables and its derivatives. This uncertainty information is propagated to the SSRE by weighting each term, where weights are inversely proportional to the variance of the terms.

\subsection{Physics Coupling}
Based on above analysis, we describe the ST-PCNN to model the heterogeneous properties of spatio-temporal data and to reveal the homogeneous physics from the data. Here, a stacking coupling mechanism is proposed to integrate the obtained physics into the spatio-temporal learning. As shown in Figure \ref{fig:connection}, at each time step $t$ and location $i,j$, the FN produces the initial prediction $\hat{\mathcal{S}}_{he}^{(t+1,i,j)}$ based on the current observation $\mathcal{S}^{(t,i,j)}$, the hidden states $h^{(t-1,i,j)}$ from previous-step (within LSTM), and the lateral info from its neighbors. The \textit{heterogeneous} initial prediction leverages its own specific local attributes only. Regarding the integration of PDE function learnt from the PN, the previous-step initial prediction $\hat{\mathcal{S}}^{(t-1,i,j)}$ and the current observations $\mathcal{S}^{(t,i,j)}$ are fed into the learnt PDE to derive the numerical solution $\hat{\mathcal{S}}_{ho}^{(t+1,i,j)}$, which is the \textit{homogeneous} of the dynamics regularized by governing physics. Finally, a coupling layer with parameters $\theta_{\mathcal{C}}=\left[\mathcal{W}_{\mathcal{C}}, b_{\mathcal{C}}\right]$ is used to produce the final prediction $\hat{\mathcal{S}}^{(t+1, i,j)}$ by synthesizing $\hat{\mathcal{S}}^{(t+1,i,j)}_{he}$ and $\hat{\mathcal{S}}_{ho}^{(t+1,i,j)}$ as:
\begin{equation}\label{eq:coupling}
    \hat{\mathcal{S}}^{(t+1,i,j)} = Relu([\hat{\mathcal{S}}^{(t+1,i,j)}_{he},\hat{\mathcal{S}}_{ho}^{(t+1,i,j)}]\mathcal{W}_{\mathcal{C}}^T + b_{\mathcal{C}})
\end{equation}

In this paper, the ST-PCNN training is presented in Algorithm \ref{algorithm:ST-PCNN} with supervised loss including sum of $l_1$-norm and $l_2$-norm loss.

\begin{algorithm}[t]
{\small
\SetAlgoLined
\SetKwInOut{Input}{Input}
\SetKwInOut{Output}{Output}
\SetKwBlock{Initialize}{Initialize}{}
\Input{$\mathbb{D}_{temp} = \{(\bm{x}, t, \mathcal{S})|\bm{x}\in \Omega^n_{temp}, t\in T_w, \mathcal{S}\in\mathbb{R}\}$\;}
\Initialize{
Neural Network parameters: $\theta_{\mathcal{F}},\theta_{\mathcal{C}}$\;
Static info: $\mathbf{\vec{p}} \leftarrow \textup{Positional-Encoding}(d,\Omega)$\;
Lateral info: $\L\leftarrow\mathbf{0}$\;}
\For{number of epochs}{
\For{$t$ in $T_w$}{
\For{$i,j$ in $\Omega_{temp}^n$}{
$\left[\hat{\mathcal{S}}_{he}^{(t+1,i,j)}, \hat{\L}^{(t+1,i,j)}\right]\leftarrow\mathcal{FN}(\mathbf{\vec{p}}^{(i,j)}, \mathcal{S}^{(t,i,j)}, \L^{(t,i,j)}, \theta_{\mathcal{F}})$\;
$\hat{\mathcal{S}}_{ho}^{(t+1,i,j)}\leftarrow{PDE}(\mathbf{\vec{p}}^{(i,j)}, \hat{\mathcal{S}}^{(t-1,i,j)}, \mathcal{S}^{(t,i,j)}, \lambda)$\;
$\hat{\mathcal{S}}^{(t+1,i,j)}=Coupling(\hat{\mathcal{S}}_{he}^{(t+1,i,j)}, \hat{\mathcal{S}}_{ho}^{(t+1,i,j)}, \theta_{\mathcal{C}})$\;
}
Update $\L^{(t+1,:)}\leftarrow \hat{\L}^{(t+1,:)}$\;}
$loss\:\ell\leftarrow \frac{1}{n}\left \| \mathcal{S}^{(t,:)}-\hat{\mathcal{S}}^{(t,:)} \right \|_1+\frac{1}{n} \left \| \mathcal{S}^{(t,:)}-\hat{\mathcal{S}}^{(t,:)} \right \|_2$\;
Update $\theta_{\mathcal{F}},\theta_{\mathcal{C}}\leftarrow loss \: \ell$.backward()}}
\caption{ST-PCNN Training}
\label{algorithm:ST-PCNN}
\end{algorithm}

\begin{table}[t]
\vspace{-3mm}
\caption{Statistics of the datasets.}
\centering
\small
\setlength\tabcolsep{3pt}
\begin{tabular}{@{}lccc@{}}
\toprule
Data Sets & \# of Time Points & Grid Size & Sampling Rate \\ \midrule[0.5pt]
Reflected Wave &8,000 &16$\times$16 &0.1s \\
LC of GoM &1,810 &29$\times$36 &12h \\ \bottomrule
\end{tabular}
\label{table: data}
\vspace{-3mm}
\end{table}

\section{Experiments}
We present two spatial-temporal prediction tasks: (1) reflected wave prediction; and (2) the Gulf of Mexico loop current prediction. Since the two tasks have different strengths of spatial-temporal correlation, as well as different scalability and noise levels, they can verify the generalizations of ST-PCNN with active learning and other baselines over different scenarios. 

Specifically, in reflected wave prediction, an flow in a position strongly corresponds to the flow in its neighbour position, which indicates a strong correlation. However, the correlation is weaker and more difficult to measure in GoM loop current prediction, where the velocity in one position may not indicates changes in other positions.

\subsection{Datasets}
The statistics of the two evaluated datasets are summarized in Table \ref{table: data} and introduced as below. 

\begin{figure}[t]
\centering
\includegraphics[width=0.35\textwidth]{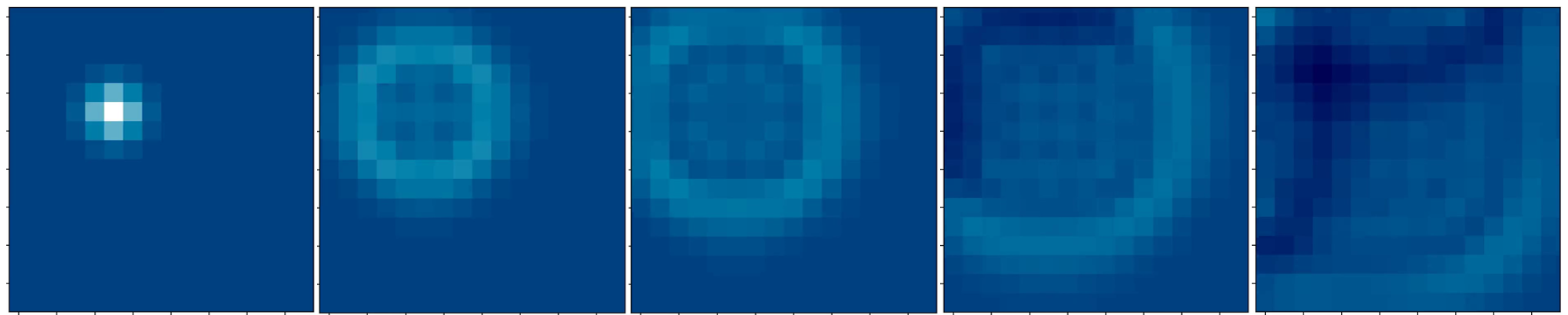}
\caption{Exemplary circular wave with reflecting borders. Plots from left to right denote temporal evolving of the circular wave (propagate from the center to boundaries with reflecting effects caused by boundary conditions.}
\label{fig:data_wave}
\end{figure}

\noindent\textbf{\textit{Reflected Wave Simulation Data}} As illustrated in Figure \ref{fig:data_wave}, single waves are propagating outwards, where waves are reflected at borders such that wave fronts become interactive. The following 2D wave equation was used for reflected wave data generation:
\begin{equation}\label{equ:wave}
    \frac{\partial ^{2}u}{\partial t^{2}} = c^2\left ( \frac{\partial ^{2}u}{\partial x^{2}} + \frac{\partial ^{2}u}{\partial y^{2}}\right)
\end{equation}
The PDE solutions to generate data were solved numerically using an explicit central difference approach:
\begin{equation}\label{Equ:second order}
\frac{\partial ^{2}u}{\partial b^{2}} = \frac{u(b+h)-2u(b)+u(b-h)}{h^2} = u_{bb} 
\end{equation}
where $b$ stands for a variable of function $u$, and $h$ is the approximation step size. In the case of calculating simulated wave data, we apply Eq. (\ref{Equ:second order}) to Eq. (\ref{equ:wave}) to obtain:
\begin{equation}
c^2(u_{xx}+u_{yy}) = \frac{u(x,y,t+\Delta t)-2u(x,y,t)+u(x,y,t-\Delta t)}{\Delta_t^2}
\end{equation}
which can be solved for $u(x,y,t+\Delta t)$ to obtain an equation for determining state of the field at the next time step $t+\Delta t$ at each point.

Both the boundary conditions (when $x < 0$ or $x > field width$, analogously for $y$) and initial condition (in time step 0) are treated as zero. The following variable choices were met: $\Delta_t=0.1$, $\Delta_x=\Delta_y=1$ and $c=3.0$. The field was initialized using a Gaussian distribution:
\begin{equation}
    u(x,y,0)=a exp\left( -\left( 
    \frac{(x-s_x)^2}{2\sigma^2_x}+\frac{(y-s_y)^2}{2\sigma^2_y}\right)\right)
\end{equation}
with amplitude factor $a=0.34$, wave width in $x$ and $y$ directions $\sigma^2_x=\sigma_y^2=0.5$, and $s_x$, $s_y$ being the starting point or center of the circular wave.

\begin{figure}[t]
  \centering
  \includegraphics[width=0.4\textwidth]{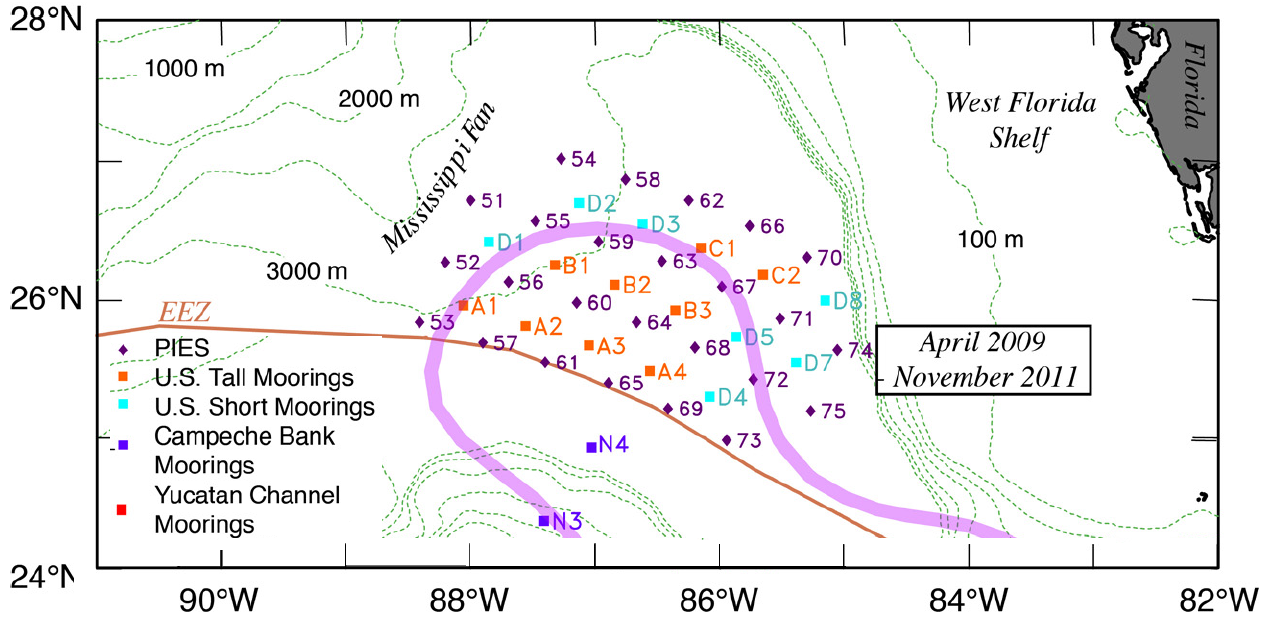}
  \caption{Locations of moorings and Pressure-Recording Inverted Echo Sounders (PIES) deployed in the U.S. and Mexican sectors in the eastern GoM.}
  \label{fig:data_location}
\end{figure}

\noindent\textbf{\textit{Gulf of Mexico (GoM) Loop Current Data}} As illustrated in Figure \ref{fig:data_location}, the sensor array are placed in the GoM region, covered from $89^o W$ to $85^o W$, and $25^o N$ to $27^o N$ with 30–50 $km$ horizontal resolution, where the Loop Current (LC) extended northward and, more importantly, where eddy shedding events occurred most often \cite{hamilton2016loop}. This sensor array consisted of 25 pressure-recording inverted echo sounders (PIES), 9 full-depth tall moorings with temperature, conductivity and velocity measurements, and 7 near bottom current meter moorings deployed under the LC region \cite{hamilton2016loop}. The data set contains velocity data gathered from June 2009 to June 2011. Since the sampling frequency from multiple sensors varied from minutes to hours, the dataset was initially processed with a fourth order Butterworth filter and sub-sampled at 12-hours intervals, leading to a total of 1,810 records (905 days). In this experiment, we further subsample the data at 7-days interval.

\subsection{Metrics \& Benchmark Models}
We evaluate the models based on the Mean Square Error (MSE). After each active training step, the performance is evaluated by single step prediction on unseen data. Our proposed model is compared by ablation study with the following methodological categories:
\begin{enumerate}
    \item \textbf{Random sampling}: At each data collection step, the positions were randomly selected from $\Omega$ at the rate of $10\%$, $20\%$ and $40\%$. This is also called passive learning, representing the widely used method in which candidate data points for training a prediction model are chosen at random.
    \item \textbf{Without physics learning}: the physics network (PN) is disabled in ST-PCNN, denoted as ST-\bcancel{PC}NN (remove PN and coupling layer). The prediction is simply produced by the forecasting network (FN).
    \item \textbf{All data available}: Suppose an optimal case that data from all positions $\Omega$ are available, i.e., $100\%$. 
\end{enumerate}

\subsection{Implementation}\label{app:hyper}
We conduct experiments on a 64-bit Ubuntu 18.04 computer with Intel 3.70GHz and 62.5GB memory, 2 NVIDIA Quadro RTX 5000 GPUs (16GB DDR6). The ST-PCNN is built on Pytorch. The hyper-parameter setting of ST-PCNN is as follow: The dimension of embedded static vector $\mathbf{\vec{p}}$ is set to 4. The FN consists of a fully-connected layer, followed by an LSTM layer with 256 hidden units, and another fully-connected layer. The PN, based on the GP model is developed by GPytorch (a Gaussian process library implemented using PyTorch). In each active step, the training epoch is set to 10, the max iteration steps in GP and in physics learning (minimizing SSRE) are both set to 200. All the data are normalized within $(0,1)$. After each active training step, for GoM Loop Current data (a single multivariate time series), the performance is evaluated on the remaining data. For the simulated reflected waves, the model is trained on a single temporal-evolving circular wave propagated from the center while evaluated on 16 circular waves propagated from randomly selected positions.

\begin{figure}[t]
  \centering
  \includegraphics[width=0.45\textwidth]{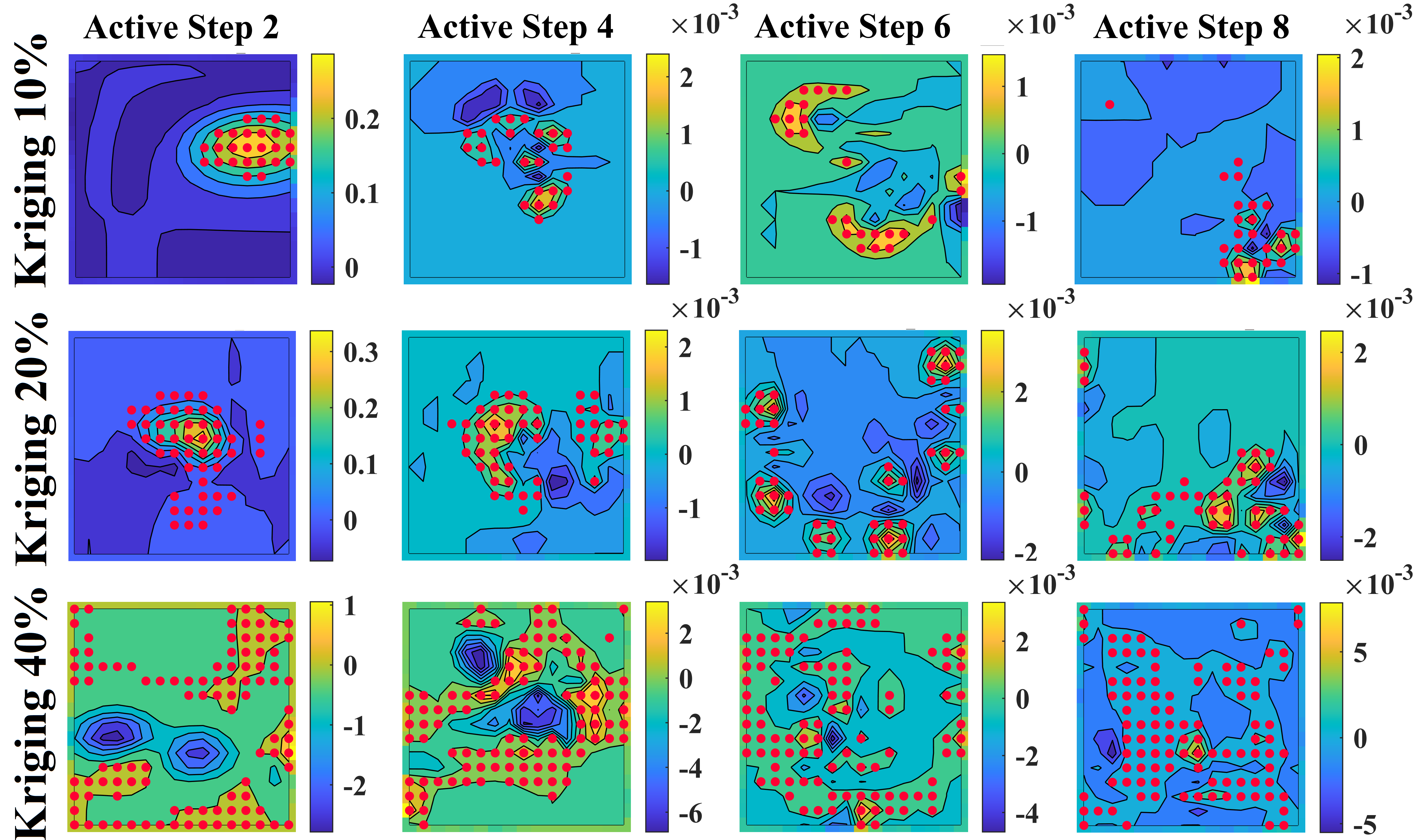}
  \caption{Example of Kriging by $\hat{s}^2$ in active learning on reflected wave.}
  \label{fig:kriging}
  \vspace{-5mm}
\end{figure}

\begin{figure}[t]
\centering
\begin{small}
\begin{subfigure}{0.48\textwidth}
\centering
\includegraphics[width=\linewidth]{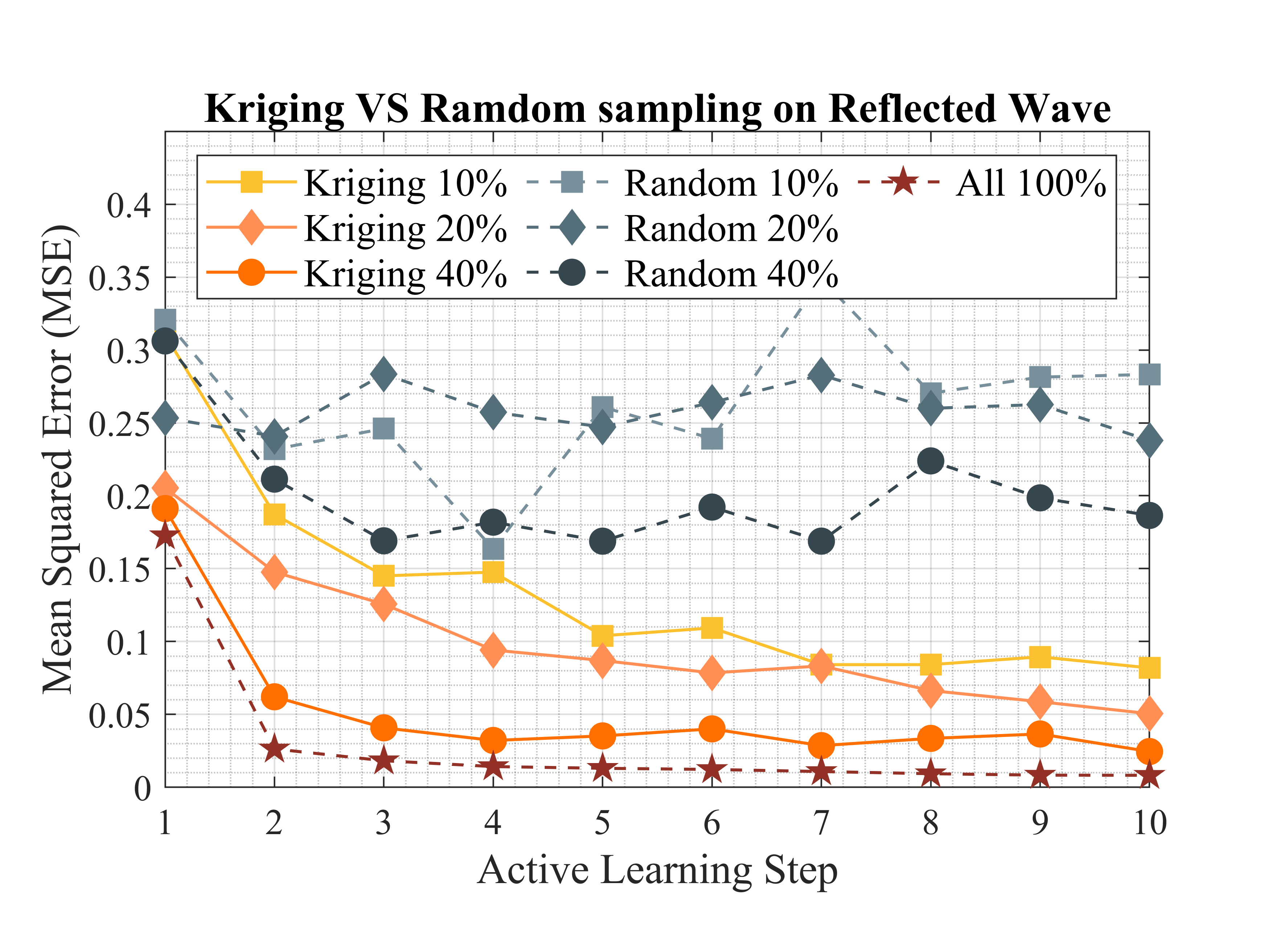} 
\end{subfigure}
\vspace{-5mm}
\begin{subfigure}{0.48\textwidth}
\centering
\includegraphics[width=\linewidth]{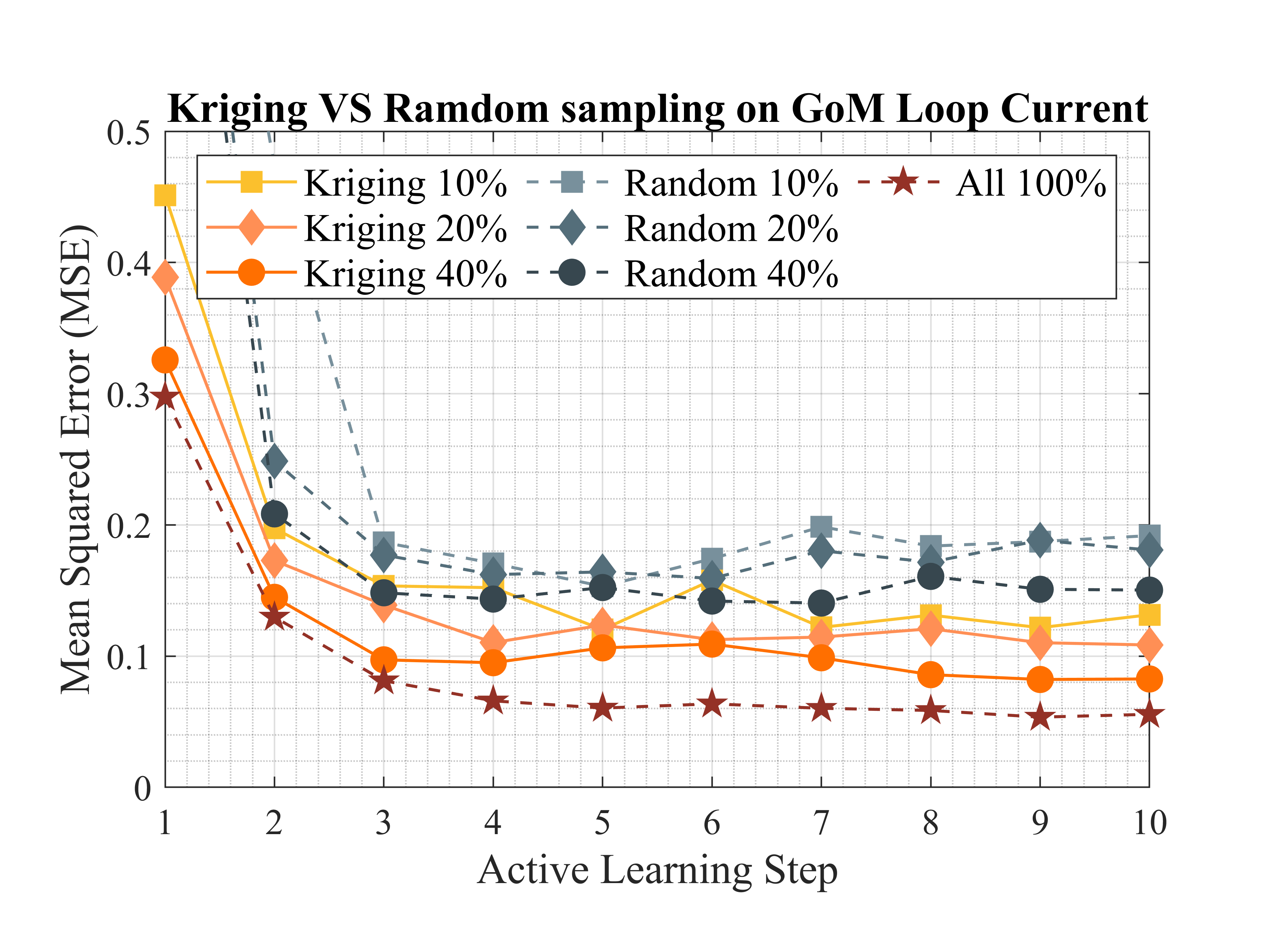}
\end{subfigure}
\caption{Explore Kriging and Random sampling effect on the performance of ST-\bcancel{PC}NN on unseen data.}
\label{fig:mse_nophy_k_vs_r}
\end{small}
\end{figure}

\subsection{Results}
\subsubsection{Kriging Sampling for Active Learning Results:}
In this section, we compare the effect of Kriging sampling-based active learning with random sampling. Our goal is to demonstrate whether the proposed active learning approach can effectively identify informative nodes/points and query their values to improve prediction accuracy.

At each data collection period (active learning step), we use Algorithm \ref{algorithm:active} to perform Kriging by using new observations at $\Omega_{temp}^n$ where $\hat{s}^2$ has maximum, which is estimated by the previous step. Figure \ref{fig:kriging} presents exemplary selected locations ($\Omega_{temp}^n$) by Kriging during active learning. As we can see, the selected locations are expended as the wave flows towards the boundary. The Kriging helps to identify those representative locations to forecast the dynamics as a whole.

In order to understand whether physics learning can indeed help improve the prediction (in combination with the active learning), we carry out an ablation study to understand the interplay between physics network (PN) and active learning, by disabling and enabling PN respectively, and observe the algorithm performance.

In the first experiment, we disable the PN (i.e., ST-\bcancel{PC}NN) to observe how sampling affects FN performance. As shown in Figure \ref{fig:mse_nophy_k_vs_r}, Kriging outperforms the random sampling in terms of both convergence and MSE. In reflected wave, ST-\bcancel{PC}NN with only $10\%$ and $20\%$ sampling does not guarantee convergence in performance, while it converges when the sampling rate rises to $40\%$. Noticeably, ST-\bcancel{PC}NN with Kriging presents a faster convergence and better prediction performance on unseen data with lower MSE. It is worth mentioning that when with a sampling rate of $40\%$, the performance is close to the optimal case where all data are observable.

\begin{figure}[t]
  \centering
  \includegraphics[width=0.45\textwidth]{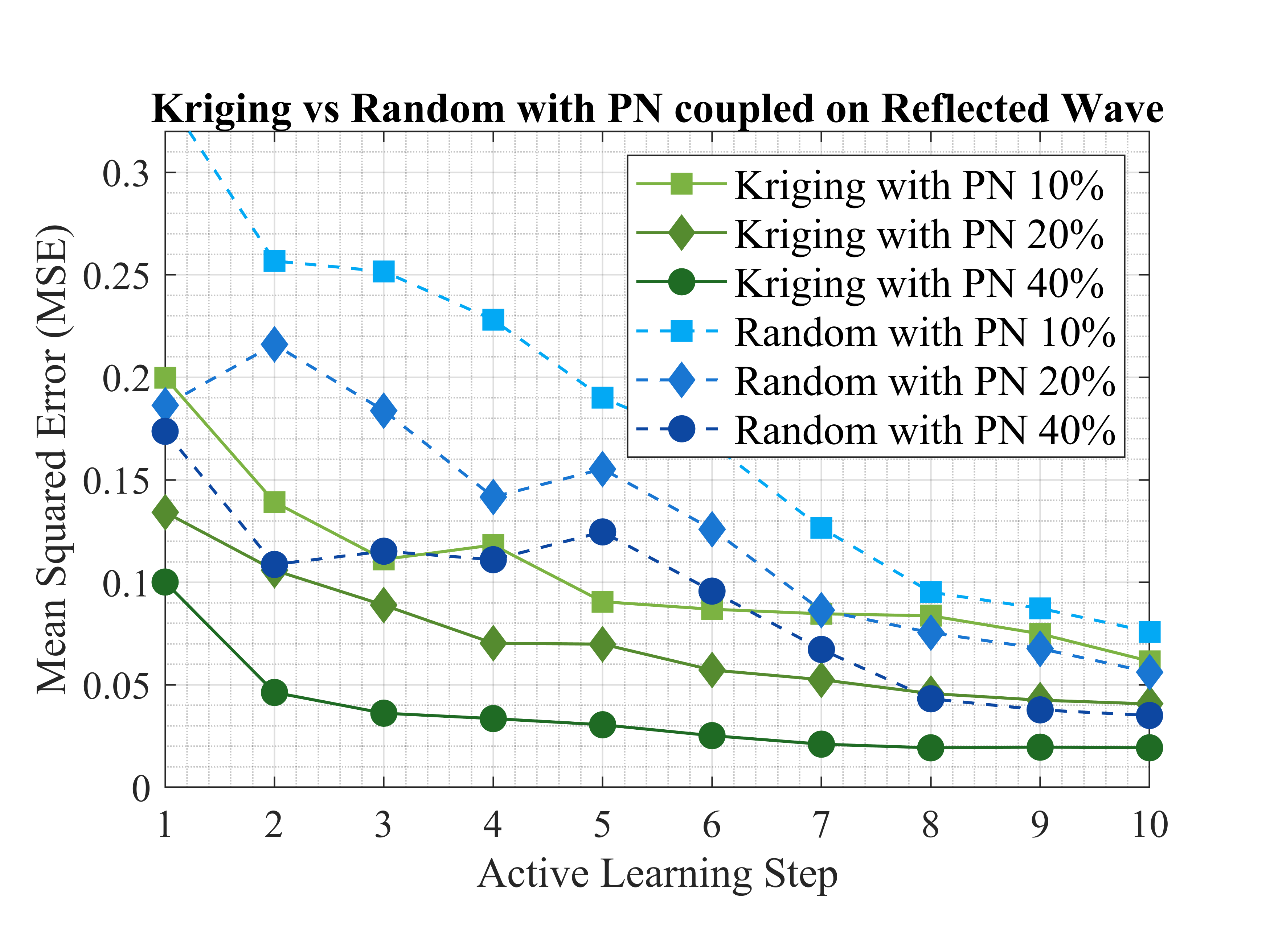}
  \caption{Explore Kriging and random sampling effect on performance of ST-PCNN on unseen data (the effect on GoM Loop Current refers to Figures \ref{fig:mse_nophy_k_vs_r} and \ref{fig:mse_physics}).}
  \label{fig:mse_wave_phy_k_vs_r}
  \vspace{-5mm}
\end{figure}

In the second experiment, we enable the PN to observe how Kriging gives credits to ST-PCNN. As shown in Figure \ref{fig:mse_wave_phy_k_vs_r}, with the benefits of PN, both Kriging and random sampling present good convergence. The benefit of Kriging is significant in pre-mid active learning. However, the performance of random sampling getting closer to Kriging as active learning goes on, which mainly due to the fact that PN gradually uncovers the underlying hidden physics from observation and thus, is able to assist the spatio-temporal networks to capture the dynamics of reflected waves.

When comparing Figure \ref{fig:mse_nophy_k_vs_r} (top panel) and Figure \ref{fig:mse_wave_phy_k_vs_r}, we can easily conclude the benefit of using PN for prediction. For example, for the same sampling rate (such as 10\%), the MSE after one active learning step is 0.3 in Figure \ref{fig:mse_nophy_k_vs_r} (without PN), where in Figure  \ref{fig:mse_wave_phy_k_vs_r}, the MSE is about 0.2. This means that PN helps the model gain more accurate predictions. After 10 active learning steps, the MSE without PN is about 0.09, whereas the one using PN is about 0.06.

\begin{figure*}[t]
\centering
\begin{subfigure}{0.465\textwidth}
\centering
\includegraphics[width=\linewidth]{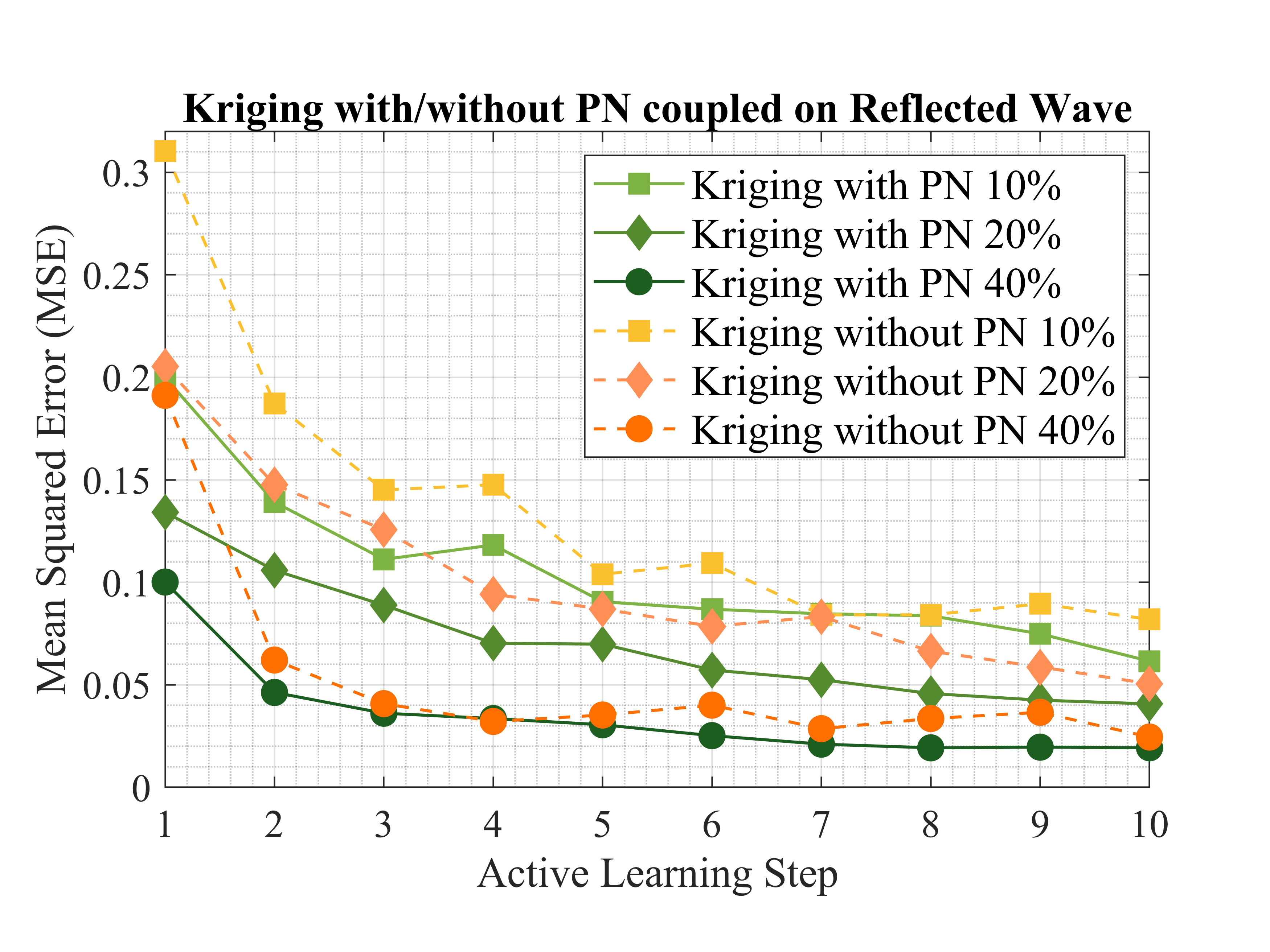} 
\end{subfigure}
\begin{subfigure}{0.465\textwidth}
\centering
\includegraphics[width=\linewidth]{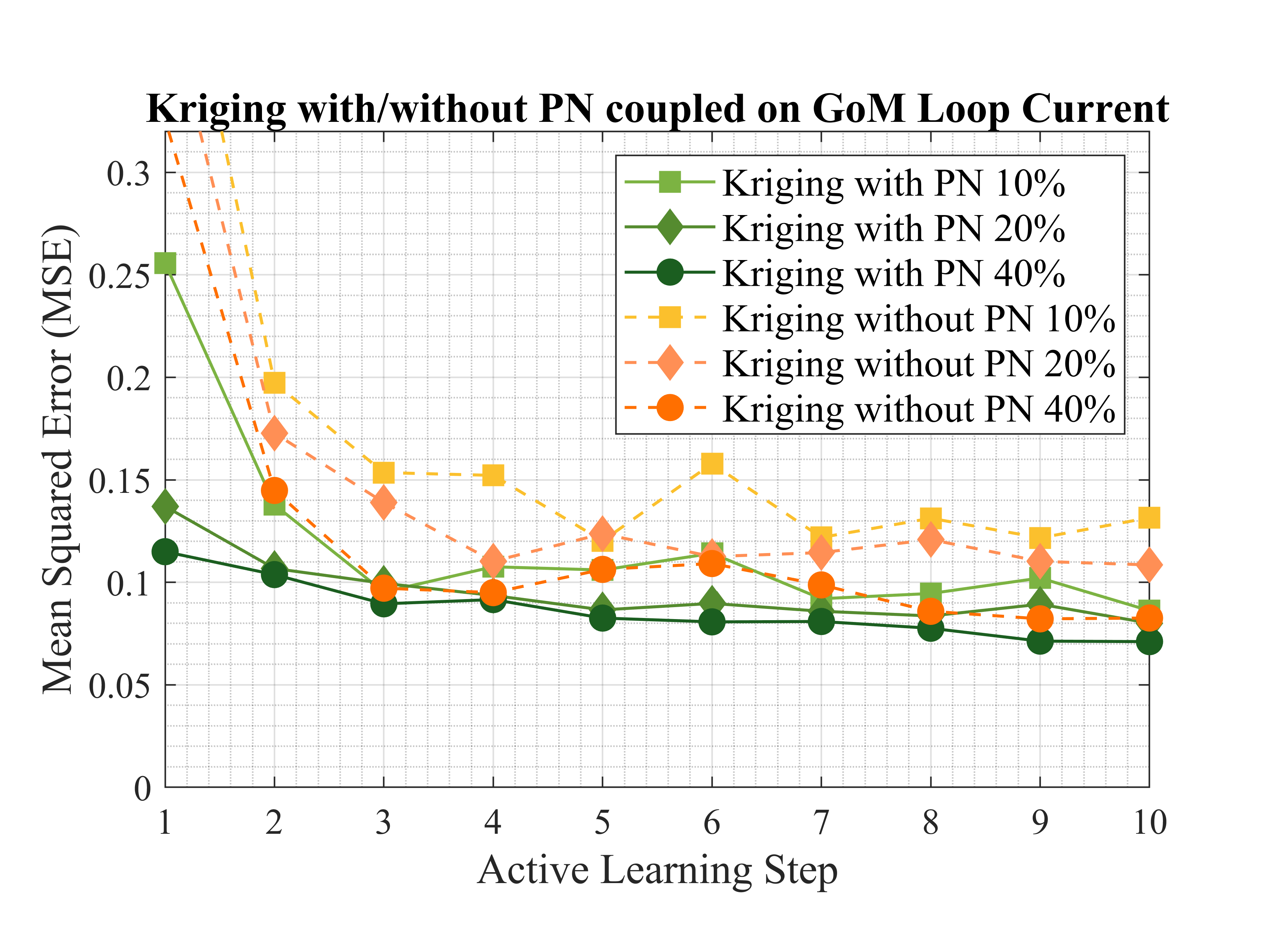}
\end{subfigure}
\caption{Explore physics effect on performance of ST-PCNN (with Kriging) on unseen data.}
\label{fig:mse_physics}
\end{figure*}

\subsubsection{Physics Learning Results:}
In this section, we further explore the benefits of PN in active learning. To begin with, we consider the optimal case where all data are observed. In this case, the Kriging for active learning is no longer needed (because all nodes are observed). For fair comparisons, we keep the same active learning step and denote by sampling \& learning step in Table \ref{table: physics}. Because all points are observed in each step, the contribution of PN can be directly observed by including \textit{vs.} excluding PN in the learning. 

In Table \ref{table: physics}, we report the results from the above experiments. ST-PCNN$^{\blacktriangle}$ with wave equation (Eq. (\ref{equ:wave})) informed is regarded as a reference here, proving that if the governing physics is known, our model could perfectly capture the spatio-temporal dynamics. The MSE score of ST-PCNN is better than that of ST-\bcancel{PC}NN at any period, indicating that ST-PCNN (with physics learning) can capture both \textit{homogeneity} (underlying physics) and \textit{heterogeneity} (localized information) in modeling the evolution and dynamics of the observing system. 

Then, we explore how PN gives credits to the ST-PCNN model with partially observed data. In the random sampling, by comparing the grey curves in Fig. \ref{fig:mse_nophy_k_vs_r} and blue curves in Fig. \ref{fig:mse_wave_phy_k_vs_r}, PN boosts the prediction accuracy with fast convergence by uncovering the hidden physics from randomly collected data. Furthermore, from Fig. \ref{fig:mse_wave_phy_k_vs_r}, we found that random sampling approximates to Kriging as time goes on. The reason lies in that, during the active learning, as new data being appended to the existing training set at each step, the training set is growing in the time dimension (since location number is fixed). The PDE learned by PN is approaching the real one by leveraging the ever-growing dataset. The advantage of Kriging is reduced as ST-PCNN gradually captures the wave dynamics by leveraging the PDE learning. This can also be found in Fig. \ref{fig:mse_physics}, which directly shows the ST-PCNN performance with \textit{vs.} without PN, by using different sampling ratios. In Figure \ref{fig:mse_physics}, we do not observe significant differences of performance in the late stage. This observation asserts that Kriging helps ST-PCNN capture the dynamics of the system, which is obvious when data are limited, resulting in lower MSE and faster convergence in the early stage.

\begin{table}[t]
\caption{The evaluation of physics effect on performance.}
\centering
\small
\setlength\tabcolsep{3pt}
\begin{tabular}{lllccccc}
\toprule
\multirow{2}{*}{Data} & \multirow{2}{*}{Model} & \multirow{2}{*}{Physics} & \multicolumn{5}{c}{Sampling \& Learning Step} \\\cmidrule(l){4-8} &  &  & 2 & 4 & 6 & 8 & 10 \\ \midrule[0.5pt]
\multirow{3}{*}{Wave}
& ST-PCNN$^{\blacktriangle}$ & Eq. (\ref{equ:wave}) & 0.0075 & 0.0034 & 0.0019 & 8.73$e^{-5}$ & 5.27$e^{-6}$ \\
& ST-PCNN & self-learnt & 0.0192 & 0.0088 & 0.0083 & 0.0066 & 0.0057\\
& ST-\bcancel{PC}NN & - & 0.0263 & 0.0142 & 0.0122 & 0.0093 & 0.0082\\\hline
\multirow{2}{*}{GoM}
& ST-PCNN & self-learnt & 0.0802 & 0.0622 & 0.0410 & 0.0434 & 0.0420 \\
& ST-\bcancel{PC}NN & - & 0.1298 & 0.0659 & 0.0636 & 0.0586 & 0.0556\\
\bottomrule
\end{tabular}
\label{table: physics}
\end{table}

\section{Related Work}
A rising question for spatio-temporal prediction is where and when to query the data to build accurate predictive model staying within a maximum budget for data collection, while the pool of candidates is very large and/or unequally distributed \cite{aryandoust2020active}. Active learning (AL) provides solutions to this question. AL is well-studied in data-driven based classification issue \cite{faghihpirayesh2020motor,mondal2020alex,tsiakmaki2020fuzzy,meduri2020comprehensive}, where vast amounts of unlabeled data is available, whereas labeled data is scarce (since the labeling process is time-consuming and costly). A first category of AL based on informativeness that exploits the uncertainty of the classifier predictions to select informative samples \cite{fu2013survey}, and the uncertainty are commonly estimated by least confidence first \cite{agrawal2021active,culotta2005reducing}, margin sampling \cite{pereira2019empirical,scheffer2001mining,balcan2007margin}, or entropy \cite{siddiqui2020viewal,qiu2016maximum}. A second category is based on density that aims to enhance the representativeness of the selected samples \cite{mondal2020alex,settles2008analysis,sener2017active, li2012incorporating,huang2010active,meduri2020comprehensive}. 

However, none of these works explore or being applicable to the spatio-temporal modeling field. Recently, \cite{aryandoust2020active} introduces an AL approach for electricity demand forecasting based on a novel \textit{embedding entropy} metric for querying candidate data. Notably, AL is also explored in solving PDEs. \cite{arthurs2020active} proposes AL algorithm for training neural network to predict PDE solutions over entire parameter space, using training data from a minimal region. In \cite{raissi2017inferring}, AL is used to quantify uncertainty in resulting predictive posterior distributions and naturally lead to adaptive solution refinement, where the only observables are scarce and noisy multi-fidelity data. \cite{zhang2019quantifying} demonstrate that dropout can quantify the uncertainty of deep neural networks in solving forward and inverse differential equations and serves as useful guidance for active learning.

\section{Conclusion}

In this paper, we proposed a physics-coupled active learning framework for the accurate prediction of spatio-temporal dynamical systems. The essential goal is to query only a small subset of the observable locations to learn the physics underlying the system and predict future values of the system with minimum errors. We argued that real-world observing systems are often challenged by \textit{heterogeneity} and \textit{homogeneity}. The proposed framework, ST-PCNN, consists of two neural networks, a forecasting network (FN) and a physics network (PN). The FN is learned spatially across all locations, allowing interaction with neighbors, to learn local \textit{heterogeneity}. The PN, on the other hand, is a Gaussian process-based model to uncover \textit{homogeneity} of the system (i.e., estimating the physics parameters). Active learning with Kriging is employed to identify locations for representative observations that minimize the prediction error and reduce MSE with scarce data and query constraints. We validated ST-PCNN on both synthetic and real-world datasets. The results show that ST-PCNN can provide accurate spatio-temporal predictions for dynamical systems, by querying only a small portion (40\% or less) of observable points.

\begin{acks}
This work was supported in part by the U.S. National Academy of Sciences Gulf Research Program and the U. S. National Science Foundation through Grant Nos. OAC-2017597 and IIS-1763452.
\end{acks}

\bibliographystyle{ACM-Reference-Format}
\bibliography{acmart}

\end{document}